\documentclass[conference]{IEEEtran}
\IEEEoverridecommandlockouts

\usepackage{cite}
\usepackage{amsmath,amssymb,amsfonts}
\usepackage{algorithmic}
\usepackage{graphicx}
\usepackage{textcomp}
\usepackage{xcolor}

\def\BibTeX{{\rm B\kern-.05em{\sc i\kern-.025em b}\kern-.08em
    T\kern-.1667em\lower.7ex\hbox{E}\kern-.125emX}}

\makeatletter
\newcommand{\linebreakand}{%
  \end{@IEEEauthorhalign}
  \hfill\mbox{}\par
  \mbox{}\hfill\begin{@IEEEauthorhalign}
}
\makeatother
\begin{document}

\begin{titlepage}

\end{titlepage}

\title{Towards Interpretable and Trustworthy Time Series Reasoning: A BlueSky Vision\\
\thanks{\textsuperscript{*}Corresponding authors}
}

\author{\IEEEauthorblockN{Kanghui Ning}
\IEEEauthorblockA{\textit{School of Computing} \\
\textit{University of Connecticut}\\
Storrs, CT \\
kanghui.ning@uconn.edu}
\and
\IEEEauthorblockN{Zijie Pan}
\IEEEauthorblockA{\textit{School of Computing} \\
\textit{University of Connecticut}\\
Storrs, CT \\
zijie.pan@uconn.edu}
\and
\IEEEauthorblockN{Yushan Jiang}
\IEEEauthorblockA{\textit{School of Computing} \\
\textit{University of Connecticut}\\
Storrs, CT \\
yushan.jiang@uconn.edu}
\linebreakand
\IEEEauthorblockN{Anderson Schneider}
\IEEEauthorblockA{\textit{Department of Machine Learning Research} \\
\textit{Morgan Stanley}\\
New York, NY \\
anderson.schneider@morganstanley.com}
\and
\IEEEauthorblockN{Yuriy Nevmyvaka\textsuperscript{*}}
\IEEEauthorblockA{\textit{Department of Machine Learning Research} \\
\textit{Morgan Stanley}\\
New York, NY \\
yuriy.nevmyvaka@morganstanley.com}
\and 
\IEEEauthorblockN{Dongjin Song\textsuperscript{*}}
\IEEEauthorblockA{\textit{School of Computing} \\
\textit{University of Connecticut}\\
Storrs, CT \\
dongjin.song@uconn.edu}
}

\maketitle

\begin{abstract}
Time series reasoning is emerging as the next frontier in temporal analysis, aiming to move beyond pattern recognition towards explicit, interpretable, and trustworthy inference. This paper presents a BlueSky vision built on two complementary directions. One builds robust foundations for time series reasoning, centered on comprehensive temporal understanding, structured multi-step reasoning, and faithful evaluation frameworks. The other advances system-level reasoning, moving beyond language-only explanations by incorporating multi-agent collaboration, multi-modal context, and retrieval-augmented approaches. Together, these directions outline a flexible and extensible framework for advancing time series reasoning, aiming to deliver interpretable and trustworthy temporal intelligence across diverse domains.
\end{abstract}


\section{Introduction}

Time series analysis has long been a cornerstone of data mining, supporting decision-making in domains such as healthcare~\cite{jin2018treatment,zhang2020adversarial,feuerriegel2024causal}, finance~\cite{zhang2017stock,rezaei2021stock}, energy~\cite{tzelepi2023deeplearningenergytimeseries}, climate science~\cite{sun2021solar,chen2022physics,chen2023physics}, and transportation~\cite{guo2019attention,zhang2021traffic,ji2023spatio}. While decades of effort and progress have been made in statistical modeling~\cite{Anderson1976TimeSeries2E,taylor2018forecasting} and deep learning~\cite{qin2017dual,lai2018modeling} towards increasingly accurate predictions, these methods remain largely rooted in pattern recognition, \textit{i.e.}, capturing temporal patterns in the time series data to produce outcomes without providing an explicit reasoning process. As real-world applications often not only require accurate prediction but also need desired interpretability, reliability, and trustworthiness, especially for mission critical scenarios, a new paradigm, \textit{i.e.}, time series reasoning is becoming increasingly important and popular. Below, we provide a formal definition of time series reasoning, discuss why this new paradigm is urgently needed, and outline our BlueSky vision to advance this field.

\subsection{What is Time Series Reasoning?}

Time series reasoning refers to the process by which a model or system reasons over temporal data to provide rationales and address tasks such as forecasting, classification, and question answering jointly.
Let $\mathcal{X} = \{x_t\}_{t=1}^T$ denote a temporal data sequence, where each $x_t \in \mathbb{R}^d$ represents a $d$ dimensional observation at time $t$, which could be multivariate or multi-modal. Let $\mathcal{C}$ denote optional contextual or auxiliary information, such as metadata, textual descriptions, or external signals.

\textbf{Time series reasoning} denotes the process in which a model 
$f_\theta$ generates an output $y \in \mathcal{Y}$ for a task $\mathcal{T}$ where $\theta$ denotes model parameters, 
along with an intermediate reasoning path $\mathcal{R}$ that connects the 
input and context to the final outcome. Formally, this problem can be formulated as
\[
    (y, \mathcal{R}) = f_\theta(\mathcal{X}, \mathcal{C}; \mathcal{T}),
\]
where $\mathcal{R} = \{r_1, r_2, \dots, r_K\}$ denotes the sequence of $K$
reasoning steps, and 
$\mathcal{T}$ can represent a wide range of tasks, including forecasting, 
classification, question answering, anomaly detection, causal inference, \textit{etc}.

Note that this process goes beyond conventional time series inference, which typically focuses on statistical pattern recognition and direct mapping from inputs to outputs. In contrast, time series reasoning aims to interpret temporal dependencies, reveal latent structures, and perform deliberate multi-step inference to facilitate underlying tasks. It could purely operate in latent space or produce explicit human-understandable outputs. While the latter is not strictly required, providing interpretable reasoning paths or explanations is often beneficial for transparency and trustworthiness in real-world applications, especially for mission critical systems.

\begin{figure*}[t]
    \centering
    \includegraphics[width=1\linewidth]{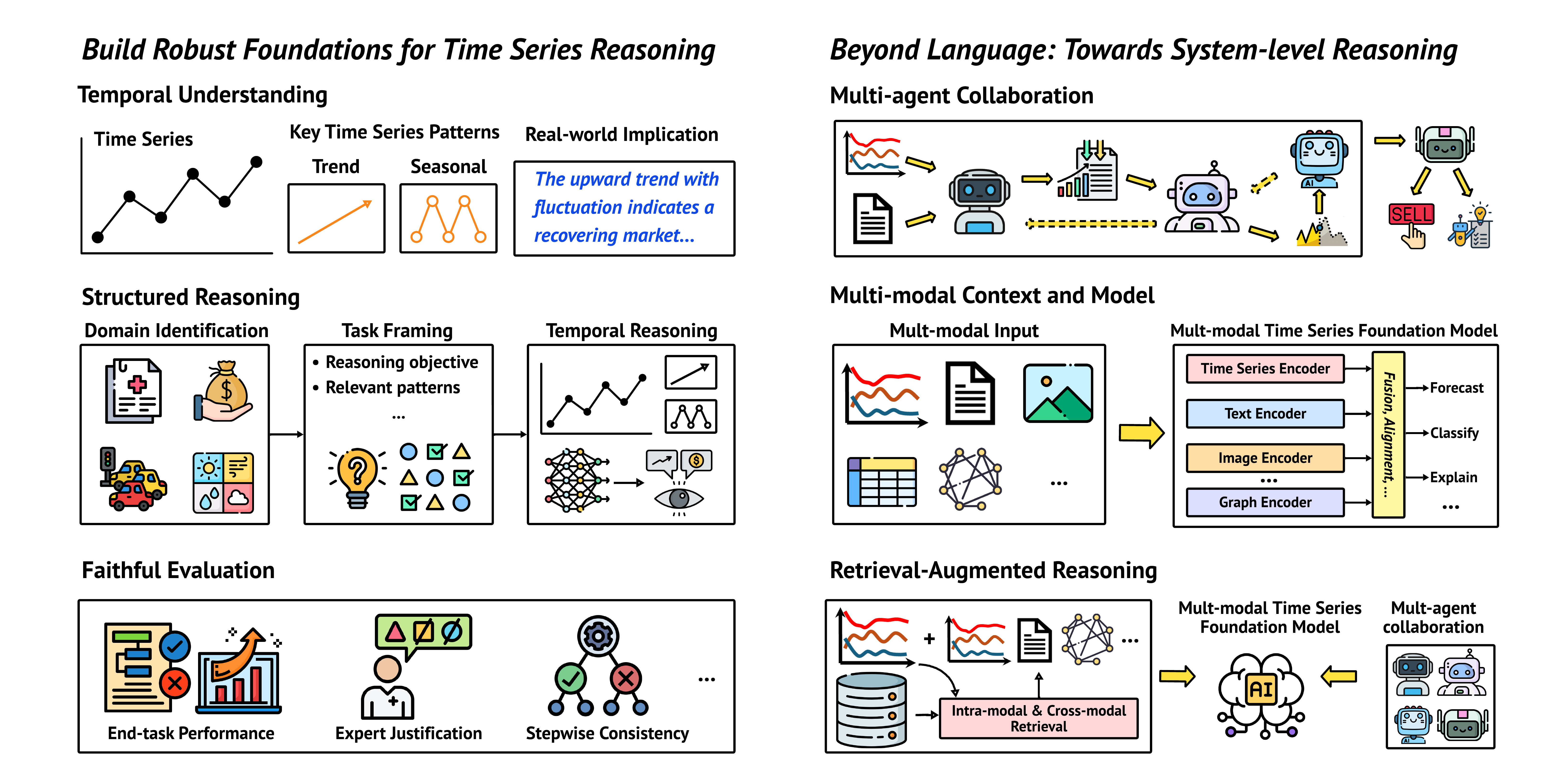}
    \caption{Overview of the proposed BlueSky idea. Left: build robust foundations for time series reasoning, including temporal understanding, structured reasoning, and faithful evaluation. Right: extend beyond language, towards system-level time series reasoning, such as multi-agent collaboration, multi-modal context and model, and retrieval-augmented reasoning.}
    \label{fig:overview}
    \vspace{-3mm}
\end{figure*}

\subsection{Why Time Series Reasoning?}

Recently, there has been a growing interest in enabling models to reason explicitly~\cite{openai2024reasoning}, moving beyond pattern recognition towards deeper understanding and better interpretability. A similar trend also emerges in time series analysis. Specifically, while traditional approaches~\cite{pan2024s,nietime,jin2024timellm,2023zeng_are_transformers_effective,wutimesnet,Zhou2022FEDformerFE,2021Wu_autoformer,zhou2021informer,liu2022non,sun2025ppgf} have demonstrated their success in capturing trends and seasonality, they often function as black boxes, producing predictions without revealing the underlying rationale. In contrast, time series reasoning aims to provide a deliberate and transparent process that can achieve desired interpretability and trustworthiness.

In particular, this is a pivotal moment to explore and advance time series reasoning, as three key factors jointly create a unique opportunity for the community. First, \textbf{tasks} in time series analysis are becoming increasingly diverse, spanning from traditional objectives such as forecasting, classification, and anomaly detection, to reasoning-intensive challenges that remain underexplored, including temporal question answering, counterfactual reasoning, cross-modal generation tasks (\textit{e.g.}, generating time series from text), \textit{etc}. These challenges emerge across diverse domains such as healthcare, finance, climate science, and industrial monitoring. Second, achieving more advanced time series analysis requires the integration of multi-modal and multi-source \textbf{data}. In time series forecasting, for example, the quality of predictions often depends on the richness of available context information. Incorporating additional contextual signals such as textual descriptions, metadata, or covariates could significantly improve performance. Recently, the emergence of several publicly available datasets and benchmarks further enables time series reasoning over heterogeneous inputs~\cite{chen2025mtbench,yang2025timeratimeseriesreasoning,kong2025timemqatimeseriesmultitask}. Finally, recent advances in \textbf{model} architectures empower time series reasoning and make it feasible. The rise of Large Language Models (LLMs) and reasoning-oriented designs, such as slow-thinking LLMs~\cite{deepseekai2025deepseekr1incentivizingreasoningcapability}, provides the capacity for deliberate, step-by-step reasoning as well as human-understandable explanations. Overall, the growing diversity of tasks, the increasing availability of multi-modal and multi-source data, and the rapid progress of reasoning-capable models create both the necessity and the opportunity to significantly advance time series reasoning.

\vspace{-2mm}
\subsection{Our BlueSky Idea}

With the opportunity, we propose a BlueSky vision focused on two complementary directions, as illustrated in Figure~\ref{fig:overview}.

\begin{itemize}
    \item \textbf{Build Robust Foundations for Time Series Reasoning:}
    Establishing the core capabilities needed for advanced reasoning, 
    including robust temporal understanding, structured multi-step reasoning, and faithful evaluation protocols. These foundations enable more transparent and trustworthy reasoning processes.
    
    \item \textbf{Beyond Language: Towards System-Level Reasoning:} 
    Overcoming the limitations of language-only explanations by 
    integrating multi-agent collaboration, multi-modal context, and retrieval-augmented techniques, enabling more interpretable and reliable reasoning.
\end{itemize}

Together, these two directions outline a forward-looking pathway towards achieving interpretable and trustworthy time series reasoning.

\section{Build Robust Foundations for Time Series Reasoning}

To realize the vision of advanced time series reasoning, we must first establish robust foundations. Current LLM-based approaches~\cite{jiang2024enpowering,pan2024s} show promise, but still face risks such as hallucination, inconsistency, and poor interpretability across three aspects: 1) \textbf{Temporal Understanding}: enabling models to capture task requirements and the characteristics of temporal data; 2) \textbf{Structured Reasoning}: ensuring models follow structured, multi-step reasoning processes aligned with temporal logic and domain knowledge; 3) \textbf{Faithful Evaluation}: assessing not only final outcomes but also the validity and consistency of reasoning processes. Below, we discuss these directions, highlighting key challenges and opportunities in building robust foundations for time series reasoning.

\subsection{Enabling Temporal Understanding} 
The first essential step in time series reasoning is understanding, which means that the model should be able to recognize task-specific requirements and accurately identify key temporal patterns and characteristics, such as anomalies, trends, periodicity, and duration.  In traditional time series analysis, temporal understanding is typically achieved through task-specific training, while in LLM-based approaches, it is often guided by prompting. Although LLM-based methods offer a flexible approach to achieving general temporal understanding, current LLMs do not natively comprehend temporal information. Existing approaches attempt to address this gap either by converting time series into raw textual representations~\cite{kong2025timemqatimeseriesmultitask} via tokenization or by transforming them into visual formats~\cite{merrill2024languagemodelsstrugglezeroshot,ni2025harnessing}. However, both approaches face significant limitations. Text-based methods struggle due to their numerical tokenization strategies~\cite{gruver2024largelanguagemodelszeroshot}, and image-based methods suffer from limitations in resolution and difficulties in accurately interpreting detailed temporal patterns~\cite{xie2025chattsaligningtimeseries}. These shortcomings restrict the potential of LLMs to achieve robust temporal understanding and reliable reasoning. Thus, we argue for the urgent development of models with robust temporal understanding, supported by native temporal representations and dedicated training tasks. This includes innovations in time series tokenization, specialized data collection, temporal-aware model architectures and dedicated optimization strategies. Progress in these directions is essential for strengthening temporal understanding, which in turn forms the foundation for more advanced reasoning processes.

\subsection{Structured Reasoning}

Reasoning in time series analysis should follow a clear and transparent process rather than jumping directly from inputs to outputs. A structured reasoning process involves a sequence of intermediate steps that connect temporal characteristics and evidence to final outputs. Such steps not only make the reasoning interpretable but also ensure that each outcome is grounded in data and domain knowledge.

However, most current LLM-based approaches do not produce truly structured reasoning outputs~\cite{wang2025slowthinkingllmsreasontime}. They often generate outcomes directly from raw data or prompts without clearly defined intermediate steps, leading to unpredictable inference patterns and a high risk of hallucination. With the emergence of advanced general reasoning models such as DeepSeek-R1~\cite{deepseekai2025deepseekr1incentivizingreasoningcapability}, some recent work has attempted to enhance reasoning structure by leveraging these models' internal multi-step reasoning abilities~\cite{zhang2025timemastertrainingtimeseriesmultimodal, luo2025timeseriesforecastingreasoning}. While this approach partially mitigates the problem, it still faces notable challenges. First, these models are primarily aligned with human preferences during training, which enables them to emulate human-like thinking but limits their capacity to capture complex temporal patterns or execute specialized algorithms required for time series analysis. Second, their reasoning processes often contain redundancy and lack standardized, optimized reasoning paths, making them inefficient and difficult to evaluate systematically.

To move forward, we call for a structured multi-step reasoning process that grounds each inference in both data and domain knowledge (\textit{e.g.}, physical constraints, causal relationships). A promising path is a multi-step reasoning framework in which each stage is explicitly defined and supported by evidence. Such a process might begin with domain identification, where the model recognizes the context of the input time series and selects appropriate reasoning strategies. Next comes task framing, which specifies the reasoning objective and highlights relevant temporal features. This is followed by temporal reasoning, in which the model forms hypotheses and draws conclusions by leveraging dependencies and domain-specific dynamics. Finally, justification and verification ensure that each step is supported by both observed data and external knowledge sources. By enforcing such structured processes, future systems can provide reasoning that is not only accurate but also interpretable and trustworthy.

\subsection{Faithful Evaluation}

Faithful evaluation is essential for establishing trust and enabling the widespread adoption of time series reasoning. It should go beyond end-task performance to assess the quality of the reasoning process itself. An ideal evaluation system should assess not only the correctness of the final output but also the temporal consistency and causal validity of the reasoning process. Such evaluation is crucial to show whether a model’s reasoning is truly sound or merely appears correct.

However, most existing benchmarks~\cite{chen2025mtbench, fatemi2024testtimebenchmarkevaluating} remain limited to end-task metrics, such as forecasting accuracy in zero-shot settings, while overlooking the quality of reasoning steps. Furthermore, standardized metrics for evaluating whether reasoning steps align with temporal dependencies or domain knowledge are still missing. As a result, current evaluations act as black boxes, making it difficult to tell reliable reasoning from invalid reasoning. To bridge this gap, we call for new evaluation protocols that incorporate measures of explanation validity and plausibility, which check if a model’s justifications match expert knowledge and domain-specific dynamics. Stepwise consistency checks are also needed to confirm that each reasoning step remains logical, consistent, and supported by the data and domain knowledge.

\section{Beyond Language: Towards System-Level Time Series Reasoning}

Recent work has explored using Large Language Models (LLMs) to generate natural language explanations for time series reasoning, aiming to provide human-understandable rationales~\cite{wang2025slowthinkingllmsreasontime, zhang2025timemastertrainingtimeseriesmultimodal, luo2025timeseriesforecastingreasoning, liu2025timer1comprehensivetemporalreasoning, chow2024timeseriesreasoningllms, liu2025pictureworththousandnumbers, xie2025chattsaligningtimeseries}. However, such approaches remain limited in their ability to support trustworthy understanding, largely due to hallucination and the inherent black-box nature of LLMs. While textual rationales can describe what a model predicts and why, the underlying generation process often remains non-transparent. To achieve more interpretable and trustworthy time series reasoning, we argue for a system-level framework integrating multiple complementary techniques beyond language-only approaches.

\subsection{Multi-Agent Collaboration}

Prior work in agentic AI has shown that LLM-based agents can coordinate with other agents or external modules to achieve complex reasoning tasks more reliably~\cite{wu2025agenticreasoningstreamlinedframework, jiang2025explainablemultimodaltimeseries, shen2024exploringmultimodalintegrationtoolaugmented, lin2024decodingtimeseriesllms, lee2024timecap}. Inspired by these developments, we propose a system in which multiple 
agents, including LLMs and auxiliary agents such as 
code execution environments, 
pretrained time series foundation models~\cite{liang2024foundationmodelsfortimeseries}, and web search services, collaborate to accomplish 
time series reasoning tasks. A comprehensive reasoning system should 
flexibly coordinate these agents and dynamically invoke them as needed 
to address complex temporal reasoning challenges. For example, an LLM agent may analyze the properties of a given dataset to determine the most suitable analytical method, whether a statistical model, deep learning architecture, or another specialized approach. The corresponding model can then be executed to generate predictions, after which an analytical LLM agent may evaluate the results, provide explanations, or suggest refinements. 
This division of labor enables each agent to contribute its unique capabilities, reducing the risks of hallucination while promoting interpretability and trustworthiness in time series reasoning.

Moreover, with explicit collaboration and specialization, agentic systems can provide trustworthy reasoning to bridge the gap between prediction and evidence-based decisions.

\subsection{Multi-Modal Context and Model}

Time series reasoning increasingly requires the integration of multi-modal context to achieve more accurate and meaningful analysis~\cite{kong2025position}. In many real-world applications, robust time series analysis requires leveraging not only the temporal sequence itself but also auxiliary signals such as textual descriptions, metadata, news, reports, or event logs~\cite{jiang2025multimodaltimeseriesanalysis}. For example, forecasting in healthcare benefits from combining vital signs with electronic health records and imaging~\cite{Johnson2016, ImageEEG}; financial prediction is improved by linking market data with news and analyst reports~\cite{2023-yu-financeforecasting}; and environmental analysis often integrates sensor measurements with geospatial and climate data~\cite{NEURIPS2024_Terra}. Incorporating such multi-modal context allows reasoning systems to uncover richer temporal dependencies and domain-specific patterns, thereby supporting predictive and analytical tasks across diverse real-world applications.

At the same time, the rapid development of multi-modal deep learning architectures provides the technical foundation for such integration. Recent advances in vision-language models, speech-language systems, and multi-modal foundation models demonstrate the feasibility of learning unified representations across heterogeneous modalities~\cite{deng2025emergingpropertiesunifiedmultimodal}. Extending these capabilities to time series offers the potential for reasoning systems that can jointly process numerical, textual, and visual information, enabling more comprehensive, interpretable, and trustworthy temporal intelligence. We call for future research to systematically develop multi-modal frameworks tailored to time series reasoning, ensuring that models can flexibly and reliably leverage information from diverse modalities to improve both accuracy and interpretability.

\subsection{Retrieval-Augmented Reasoning}

A further step towards system-level time series reasoning is the integration of retrieval-augmented generation (RAG)~\cite{lewis2020retrieval, ning2025TSRAG}. Retrieval-augmented reasoning equips time series systems with the ability to dynamically access external knowledge on demand. Instead of relying solely on the information encoded in model parameters or the observed context, the system actively queries dedicated databases or online knowledge sources to obtain context relevant to the task at hand. Retrieved evidence is then integrated into the reasoning process, allowing the model to ground its inferences in verifiable information rather than generating outputs from parametric memory alone. 

This capability is particularly valuable when reasoning requires context that is not directly available in the observed data, enhancing interpretability by grounding outputs in verifiable evidence. For example, an energy consumption anomaly may be better explained using holiday schedules or weather conditions; and clinical decision support may require consultation of medical guidelines or patient records.

By combining multi-agent collaboration, multi-modal context and modeling, and retrieval-augmented reasoning, we envision a system-level approach to time series reasoning that moves beyond language-only explanations towards more interpretable and trustworthy temporal intelligence. While these three directions illustrate promising pathways, system-level time series reasoning should remain flexible and extensible, capable of incorporating new techniques as the field advances.

\section{Conclusion}

The next stage of time series analysis calls for models that go beyond prediction, offering reasoning processes that are transparent and reliable. This paper has outlined a BlueSky vision for time series reasoning, highlighting the need to build robust foundations and develop system-level approaches that embed interpretability and trustworthiness as core principles.

Looking ahead, we see opportunities to design systems that combine these foundations with system-level capabilities, enabling explicit and verifiable reasoning while adapting flexibly to diverse domains. Such systems would not only improve predictive performance but also provide meaningful explanations and support responsible decision-making in real-world applications. We hope this vision inspires further exploration and development, advancing time series reasoning towards interpretable and trustworthy temporal intelligence.

\newpage


\bibliographystyle{IEEEtran}
\bibliography{main}

\end{document}